\documentclass[10pt,twocolumn,letterpaper]{article}

\usepackage{cvpr}              % To produce the CAMERA-READY version
\usepackage{tabularx}
\usepackage{array}
\newcolumntype{Y}{>{\centering\arraybackslash}X}
\newcolumntype{R}{>{\raggedleft\arraybackslash}X}

\definecolor{cvprblue}{rgb}{0.21,0.49,0.74}
\usepackage[pagebackref,breaklinks,colorlinks,allcolors=cvprblue]{hyperref}
\usepackage{multirow}
\hypersetup{colorlinks,
urlcolor=magenta,}
\usepackage[accsupp]{axessibility}

\def\paperID{8} % *** Enter the Paper ID here
\def\confName{CVPR}
\def\confYear{2026}

\title{IonMorphNet: Generalizable Learning of Ion Image Morphologies \\ for Peak Picking in Mass Spectrometry Imaging}

\author{
Philipp Weigand\textsuperscript{1,2,}\thanks{Equal contribution} \quad Niels Nawrot\textsuperscript{1,}\footnotemark[1] \quad Nikolas Ebert\textsuperscript{1} \quad Carsten Hopf\textsuperscript{1} \quad Oliver Wasenm\"uller\textsuperscript{1,2}\\[4pt]
\textsuperscript{1}Mannheim University of Applied Sciences, Germany \\
\textsuperscript{2}Faculty of Biosciences, Heidelberg University, Germany \\ 
{\tt\small \{p.weigand, n.nawrot, n.ebert, c.hopf, o.wasenmueller\}@th-mannheim.de}
}

\begin{document}
\maketitle
\begin{abstract}
Peak picking is a fundamental preprocessing step in Mass Spectrometry Imaging (MSI), where each sample is represented by hundreds to thousands of ion images. Existing approaches require careful dataset-specific hyperparameter tuning, and often fail to generalize across acquisition protocols. We introduce IonMorphNet, a spatial-structure–aware representation model for ion images that enables fully data-driven peak picking without any task-specific supervision. We curate 53 publicly available MSI datasets and define six structural classes capturing representative spatial patterns in ion images to train standard image backbones for structural pattern classification.
Once trained, IonMorphNet can assess ion images and perform peak picking without additional hyperparameter tuning. Using a ConvNeXt V2-Tiny backbone, our approach improves peak picking performance by $+7 \%$ mSCF1 compared to state-of-the-art methods across multiple datasets. Beyond peak picking, we demonstrate that spatially informed channel reduction enables a 3D CNN for patch-based tumor classification in MSI. This approach matches or exceeds pixel-wise spectral classifiers by up to $+7.3 \%$ Balanced Accuracy on three tumor classification tasks, indicating meaningful ion image selection. The source code and model weights are available at \url{https://github.com/CeMOS-IS/IonMorphNet}.
\end{abstract}    
\section{Introduction}
\label{sec:intro}

\begin{figure}[t]
  \centering
   \includegraphics[width=1.0\linewidth]{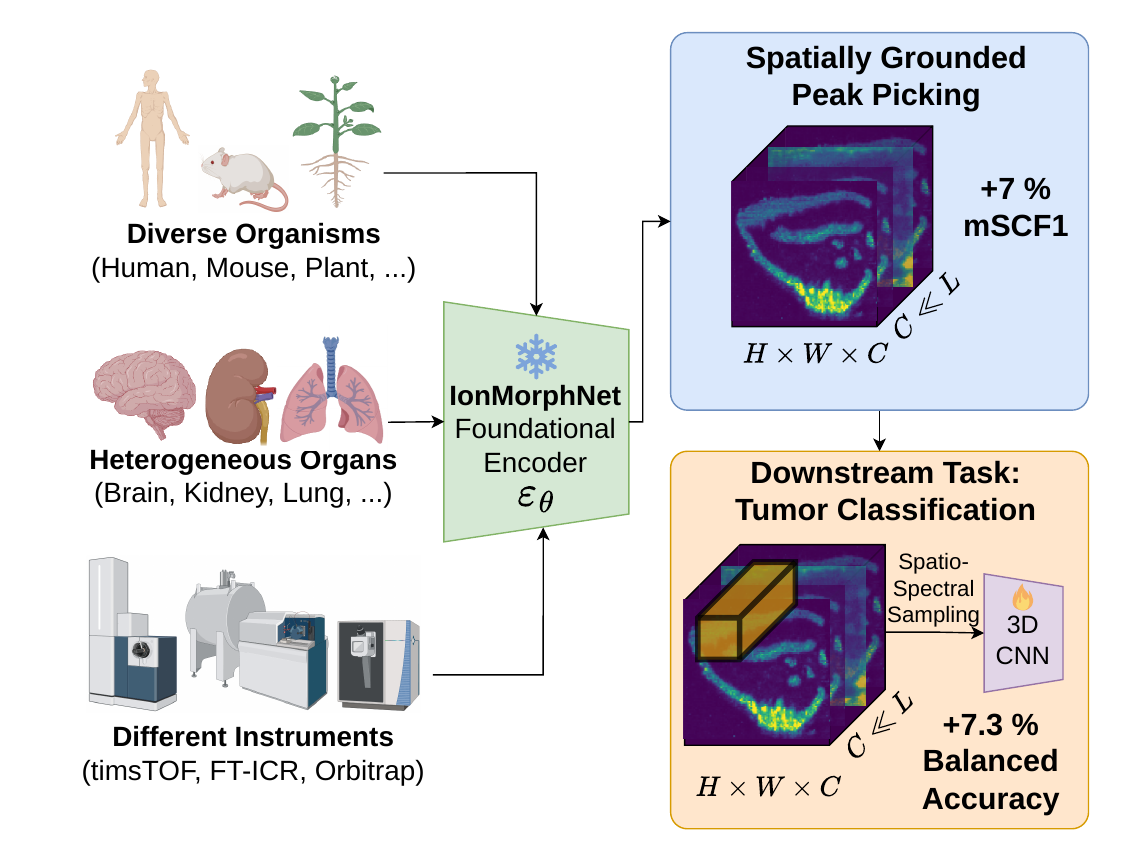}

   \caption{\textbf{Overview of IonMorphNet, a foundational ion image encoder for Mass Spectrometry Imaging (MSI).} IonMorphNet is pre-trained on morphology classification across diverse organisms, organs and instrument types. The frozen encoder performs ion image assessment, thereby enabling spatially grounded peak picking across diverse datasets. Furthermore, spatially grounded peak picking enables tumor classification with 3D CNNs which can improve Balanced Accuracy by up to $+7.3 \%$ compared to pixelwise approaches.}
   \label{fig:teaser}
\end{figure}

Mass Spectrometry Imaging (MSI) enables label-free analysis of molecular spatial distributions in biological tissues which is important for pharmaceutical research, drug development~\cite{schulz2019advanced}, discovery of candidate biomarkers~\cite{arendowski2024matrix} and histopathological tissue analysis~\cite{rittel2026cohort}. MSI produces biomolecular images with channel dimensions of up to tens of thousands of channels (m/z values). Untargeted analysis requires prefiltering of distinct signals and removal of artifacts and noise to enhance the information content in MSI data and the single channel images called \textit{ion images}. 

The standard filtering strategy is peak picking \cite{gibb2012maldiquant, bemis2015cardinal}, which is currently challenged by rule-based spatially informed approaches \cite{lieb2020peak,inglese2019sputnik, alexandrov2013testing, abu2023spatial} as well as deep-learning-based methods \cite{abdelmoula2021peak, weigand2026spatial}. However, all existing methods need hyperparameter tuning, and in the case of deep-learning-based approaches, training from scratch is necessary for each MSI dataset. This has the potential to yield favourable outcomes on particular datasets for which a method was developed. However, it can result in suboptimal generalization across other datasets, as demonstrated recently ~\cite{weigand2026spatial}. While powerful image encoders dominate the domain of natural images \cite{dosovitskiy2020image, radford2021learning, simeoni2025dinov3}, they have so far been absent from the MSI domain. Tasks like classification, segmentation, and object detection have not been widely applied in MSI due to the limited availability and high cost of annotated samples. Furthermore, the tissue sections originate from different organisms and organs, that are measured on different devices and require expert annotation to obtain labels such as tumorous and healthy tissue regions. These diverse data sources pose a challenge regarding the generalization across organisms, organs and measurement devices. Currently, most supervised tasks are performed on single datasets with their specific characteristics \cite{behrmann2018deep, inglese2017deep, abdelmoula2022massnet, kriegsmann2020mass, kanter2023classification}. Therefore, the MSI domain would greatly benefit from a generalizable ion image encoder which can be applied across various types of datasets.

In this work, we propose IonMorphNet, the first general-purpose image encoder for MSI ion images. It is designed to enhance spatially informed peak picking without any required hyperparameter tuning during inference. To enable cross-dataset generalization, we collected $53$ publicly available diverse MSI datasets covering different organisms and organs, measured with various devices, and labeled their ion images with structure-based class labels that can be transferred to any MSI dataset. This unique dataset collection enables, for the first time, large scale training of common image encoders for structure assessment of ion images. Our IonMorphNet outperforms state-of-the-art peak picking methods by $+7 \%$ Avg. mSCF1 while requiring neither retraining nor hyperparameter tuning for new datasets.
We perform ablation studies on the selected classes for peak picking, on different backbones and analyze the impact of ImageNet~\cite{deng2009imagenet} pretraining.

Furthermore, we showcase the applicability of our IonMorphNet as a preprocessing step before tumor classification. By reducing the channel dimensionality of MSI datasets, we enable the application of 3D CNNs, which exceed pixel-wise classification by up to $+7.3 \%$ Balanced Accuracy or achieve similar results.
\section{Related Work}
\label{sec:related work}
\subsection{Peak Picking}
\textbf{Classical Peak Picking.} In an MSI dataset, spectra are typically represented in either profile or centroided format. Profile spectra retain the full peak shapes and typically contain the same number of m/z values in each pixel. Centroiding, or peak picking, reduces these peak shapes to a single m/z value per peak. Most peak picking methods process spectra separately from each other, without spatial context. MALDIquant~\cite{gibb2012maldiquant} and Cardinal~\cite{bemis2015cardinal}, for instance, offer modular MSI analysis pipelines that encompass peak picking functionality. Such methods typically rely on signal-to-noise ratio thresholding applied to individual spectra to differentiate true peaks from background noise. However, they completely neglect the spatial information inherent in MSI data. Sparse frame multipliers~\cite{lieb2020peak} were proposed as the first technique to exploit both spatial and spectral (spatio-spectral) information for peak picking of structured ion images in matrix-assisted laser desorption/ionization (MALDI) time-of-flight (TOF) MSI datasets.
Unlike the methods described above, another line of work operates exclusively on ion images, disregarding the spectral dimension of MSI data. The measure of spatial chaos~\cite{alexandrov2013testing} was introduced to quantify the spatial structure of peak picked ion images, while an adaptation of the Gray Level Co-Occurrence (GCO) matrix~\cite{gadelmawla2004vision} was later proposed for analogous purposes~\cite{wijetunge2015exims}. A subsequent refinement of the spatial chaos measure was presented as part of a broader pipeline for identifying biologically meaningful molecular signals~\cite{palmer2017fdr}. It is important to note that both the spatial chaos measure and the GCO matrix approach have been specifically designed to evaluate the spatial quality of pre-selected, already peak-picked ion images, and neither constitutes a peak picking method in its own right.
SPUTNIK~\cite{inglese2019sputnik} takes a different approach, even though it similarly focuses on ion images, but is explicitly designed as a peak picking pipeline for profile MSI data. The method first constructs a dataset-level reference image via principal component analysis~\cite{pearson1901liii}, then identifies relevant peaks by correlating individual ion images against this reference.

\begin{figure*}[!t]
  \centering
   \includegraphics[width=0.85\linewidth]{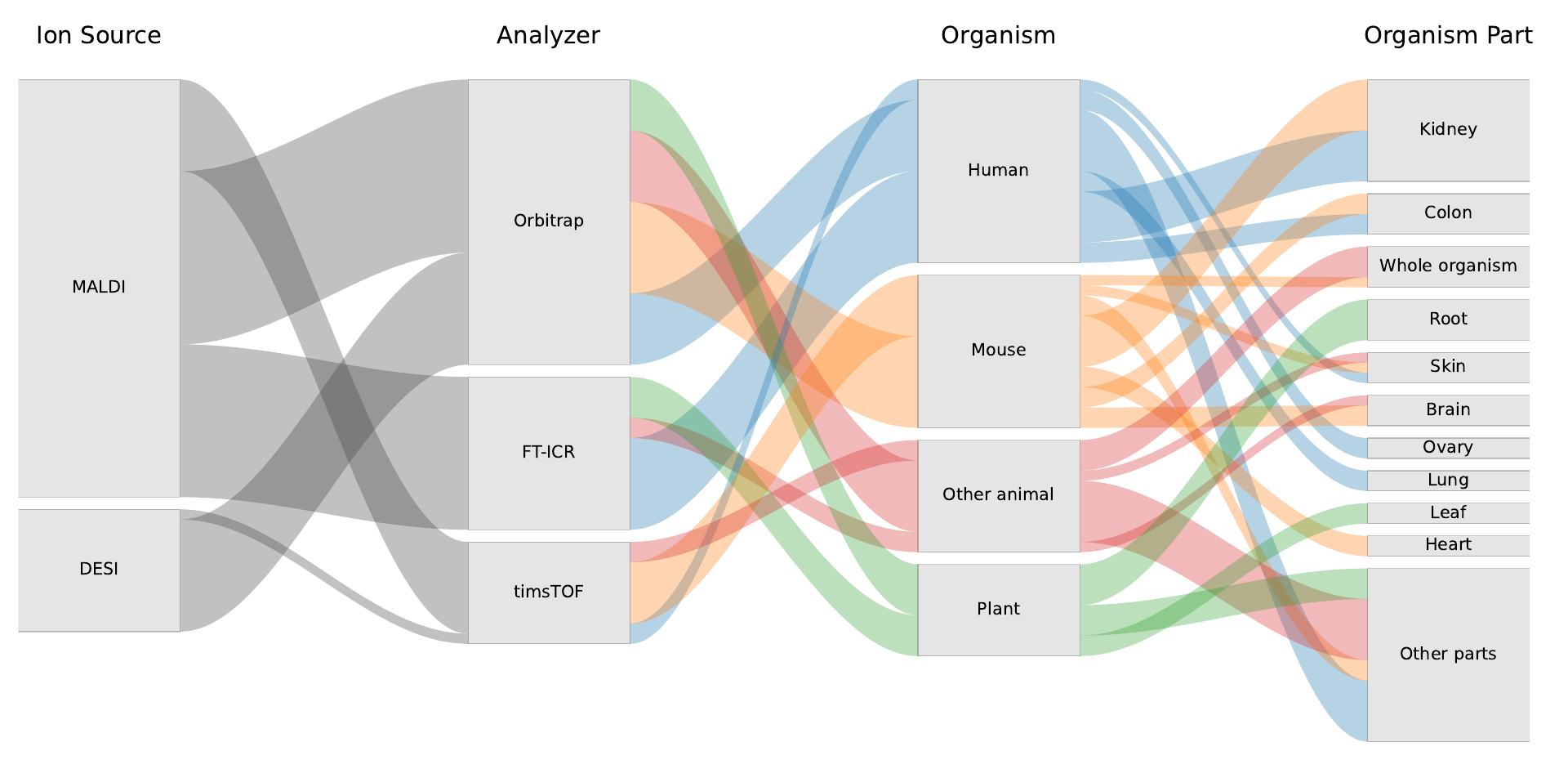}

   \caption{Overview of all $53$ collected public datasets regarding ion source, analyzer, organism and organism part.}
   \label{fig:dataset_overview}
\end{figure*}

\textbf{Deep-Learning-based Peak Picking.}
The latest peak picking approaches are both self-supervised autoencoders. msiPL~\cite{abdelmoula2021peak} utilizes a variational autoencoder~\cite{kingma2013auto} of feed-forward-layers to reconstruct single mass spectra of a MSI dataset separately. Subsequently, relevant peaks are selected by identifying the most prominent input weights of the model and the corresponding m/z-values. S\textsuperscript{3}PL~\cite{weigand2026spatial} extends this approach by processing spatio-spectral patches in a self-supervised manner using 3D Convolutions and an attention mask as the bottleneck. After training, the attention mask is analyzed for the most prominent m/z-values. However, unlike our proposed IonMorphNet, both msiPL and S\textsuperscript{3}PL have to be trained individually for each MSI dataset.

\textbf{Evaluation of Peak Picking Methods.}
Recently, the first quantitative evaluation procedure for spatially structured peak picking on profile datasets has been proposed~\cite{weigand2026spatial}. Instead of labeling every single peak in a dataset, a segmentation mask for the specific MSI dataset is correlated to all ion images to conclude a ground truth peak selection. The pearson correlation coefficient (PCC) is employed as the correlation measure between every ion image and segmented area to rank every ion image. Multiple correlation thresholds are considered to prevent a biased threshold selection and to define the metric for the performance of a peak picking method, called the mean spatial correlation F1 Score: \textit{mSCF1}. We will use the \textit{mSCF1} to evaluate our proposed IonMorphNet and compare it against the current state-of-the-art.

\subsection{Tumor Classification in Mass Spectrometry Imaging}
The spectral information in mass spectra enables the classification of different tumor types such as lung tumor~\cite{behrmann2018deep}, colorectal cancer~\cite{inglese2017deep}, bladder cancer~\cite{guo2020deep}, ductal carcinoma~\cite{kanter2023classification}, brain tumor~\cite{abdelmoula2022massnet} and renal cell carcinoma~\cite{bemis2019cardinalworkflows}. These works classify each pixel (spectrum) of an MSI measurement separately, finally leading to a segmentation map. Supervised segmentation networks, which directly output a segmentation map have not been proposed yet, as the number of available tissue samples is usually limited to two to eight tissue sections. Therefore, current approaches classify mass spectra of MSI datasets separately and report the mean balanced accuracy across all test tissue sections as their evaluation metric. Using our IonMorphNet, we will demonstrate a spatio-spectral classification that is superior to pixel-wise classification.
\begin{figure*}[t]
  \centering
   \includegraphics[width=1.0\linewidth]{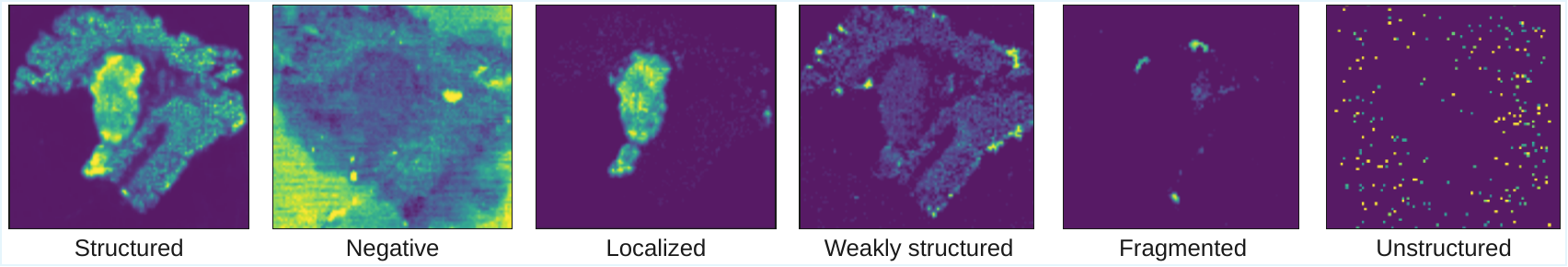}

   \caption{Overview of the six structural classes we used to label all $53$ collected MSI datasets.}
   \label{fig:overview_structural_classes}
\end{figure*}

\begin{table*}[!t]
\centering
\caption{Class distribution across training, validation and test set. Counts refer to the number of ion images per morphology class in each split; the last row reports the total sum across all splits.}
\label{tab:class_distribution}
\resizebox{\textwidth}{!}{%
\begin{tabular}{lccccccc}
\toprule
Dataset Split & Structured & Weakly Structured & Localized & Fragmented & Unstructured & Negative & Total \\
\midrule
Training (80.0 \%)    & 3,136 & 883 & 1,141 & 947 & 1,218 & 1,099 & 8,424 \\
Validation (3.5 \%) & 137  & 46  & 81   & 19  & 32   & 23   & 338  \\
Test (16.5 \%)      & 616   & 209 & 227  & 196 & 303  & 175  & 1,726 \\
\midrule
$\sum$ & 3,889       & 1,138            & 1,449      & 1,162       & 1,553         & 1,297     & 10,488 \\
\bottomrule
\end{tabular}
}
\end{table*}

\section{Method}
\subsection{Dataset Collection and Annotation}
\label{sec:dataset_and_labels}
In Mass Spectrometry Imaging (MSI), a large-scale standardized benchmark comparable to ImageNet~\cite{deng2009imagenet}, COCO~\cite{lin2014microsoft}, or CIFAR~\cite{krizhevsky2009learning} does not yet exist. MSI data are generated using a wide range of mass spectrometers, acquisition protocols, and tissue types, making cross-study standardization difficult. However, the platform METASPACE~\cite{alexandrov2019metaspace} enables researchers to upload and publicly share MSI datasets, resulting in a large and diverse collection of ion images acquired under heterogeneous experimental conditions.

In this work, we utilize 53 public datasets from METASPACE to construct a labeled dataset for learning spatial morphology in ion images. The selected datasets cover a wide range of organisms, organs, acquisition protocols, instrument types, and spatial resolutions. Although human and mouse tissue samples constitute the majority of publicly available MSI data, we intentionally limited their proportion and preferentially included datasets from plants and other animals in order to increase biological diversity. An overview of the collected datasets is shown in \cref{fig:dataset_overview}.

\paragraph{Structural Classes.}
The central task addressed in this work is the assessment of spatial structure and information content in ion images. In contrast to conventional visual recognition benchmarks such as ImageNet, MSI ion images do not contain discrete semantic objects. Instead, they represent spatial distributions of molecules whose patterns may vary in intensity, coherence, and spatial extent. Consequently, the distinction between “structured” and “unstructured” images is inherently ambiguous and subjective.

To reduce this ambiguity, we introduce a standardized annotation protocol that captures common spatial patterns observed in MSI. Rather than enforcing a binary decision, we define six structural classes that represent the most frequently occurring spatial signal patterns while remaining sufficiently general to transfer across different organisms, organs, and acquisition protocols.

The six structural classes are defined as follows: \\
\textbf{Structured:} The entire sample structure is clearly visible; signal is dense, continuous, and fills the region without gaps.\\
\textbf{Negative:} Images exhibit strong but inverted structure, the sample appears as a dark mask. \\
\textbf{Localized:} The signal is confined to specific localized regions and does not represent the full structure.\\
\textbf{Weakly Structured:} The full structure is visible but with lower intensity or thinner signal; gaps and reduced contrast may occur.\\
\textbf{Fragmented:} The signal is split into multiple distinct dense clusters; clusters are separated and not continuous. \\
\textbf{Unstructured:} The signal is random, noisy, sparse, or diffuse with no discernible pattern or anatomical correspondence. \\
A visual overview across these classes is provided in \cref{fig:overview_structural_classes}.

\paragraph{Annotation Strategy and Dataset Split.}
Using the defined structural classes, we manually annotated ion images extracted from the selected datasets. Instead of labeling all ion images within each MSI dataset, we focused primarily on images exhibiting meaningful spatial patterns. In raw MSI datasets, the majority of ion images contain little signal or are dominated by noise, which would lead to a severe class imbalance if annotated exhaustively.

To preferentially sample informative ion images, we leveraged the ranking system provided by METASPACE, which prioritizes ion signals supported by database-based molecular annotations. Ion images were generated by integrating intensities within a symmetric window around each target mass-to-charge ratio (m/z), using the dataset-specific parts-per-million (ppm) tolerance from METASPACE.

Each extracted ion image was assigned one of the six structural labels. All annotations were performed by three experts. Ambiguous cases were resolved by consensus. In total, this procedure resulted in $10,488$ labeled ion images across the $53$ datasets. To prevent leakage of spatial patterns, dataset splits are performed at dataset level. The dataset collection was split into $43$ training, $5$ validation, and $5$ test datasets. Detailed statistics on class distributions and dataset splits are provided in \cref{tab:class_distribution}. The resulting collection of labeled ion images forms the basis for training a generalizable representation model for MSI morphology.

\subsection{Ion Image Preprocessing}
Each ion image represents a single-channel spatial intensity map. Prior to training, all images undergo a standardized preprocessing pipeline consistent with the visualization procedure used in METASPACE.

First, hotspot clipping is applied to reduce the influence of sparse high-intensity spikes that commonly occur in MSI measurements. Next, each image is robustly normalized by dividing intensities by the 99th percentile value and subsequently clamped to the range $[0,1]$. This normalization ensures comparable dynamic ranges across datasets with different acquisition conditions.
After normalization, ion images are resized to a fixed resolution of $224\times224$ pixels to match the input requirements of standard image classification networks.

To improve model robustness, we apply image augmentations that preserve the underlying spatial structure. Specifically, we use random horizontal and vertical flipping as well as random rotations by multiples of $90^{\circ}$. These transformations are chosen because they do not alter the structural morphology of ion images. In contrast, region-removal augmentations such as CutMix~\cite{yun2019cutmix} or Cutout~\cite{devries2017improved} are intentionally avoided, as removing image regions may significantly alter or destroy the spatial patterns associated with a given structural label.

\subsection{Spatial Structure Learning with IonMorphNet}
Direct supervision for tasks such as MSI peak picking or ion relevance estimation is scarce and often tied to specific studies or expert assumptions. Instead of training a model directly for peak selection, we formulate spatial structure classification as a proxy learning task.

Given an ion image $I \in \mathbb{R}^{H \times W}$, the objective is to predict a structural label $y \in \{1, \dots, N\}$, where $N=6$ refers to the number of structural classes defined in \cref{sec:dataset_and_labels}. Although this task is not the final application goal, it encourages the model to learn representations that capture spatial coherence, localization, fragmentation, and noise patterns-properties that are highly relevant for downstream MSI analysis.

The proposed model, IonMorphNet, follows a standard image classification architecture consisting of an image encoder and a lightweight prediction head. The encoder maps each ion image to a fixed-dimensional embedding vector, which is subsequently passed to a classifier that predicts the structural class.

As encoders, we consider both convolutional neural networks (CNN) and vision transformer (ViT) architectures pretrained on ImageNet. Pretraining provides generic visual features that can be adapted to MSI morphology learning. The impact of ImageNet pretraining is analyzed in \cref{tab:ablation_pretraining}. After training, the model can be applied to ion images from unseen MSI datasets to assess their spatial structure.

The diversity of the curated dataset collection plays an important role in enabling the model to generalize across different biological tissues, acquisition protocols, and instrument types. By learning to discriminate between different spatial structure types across many heterogeneous datasets, IonMorphNet acquires representations that capture common morphological characteristics of MSI ion images.
\begin{figure*}[!t]
  \centering
   \includegraphics[width=1.0\linewidth]{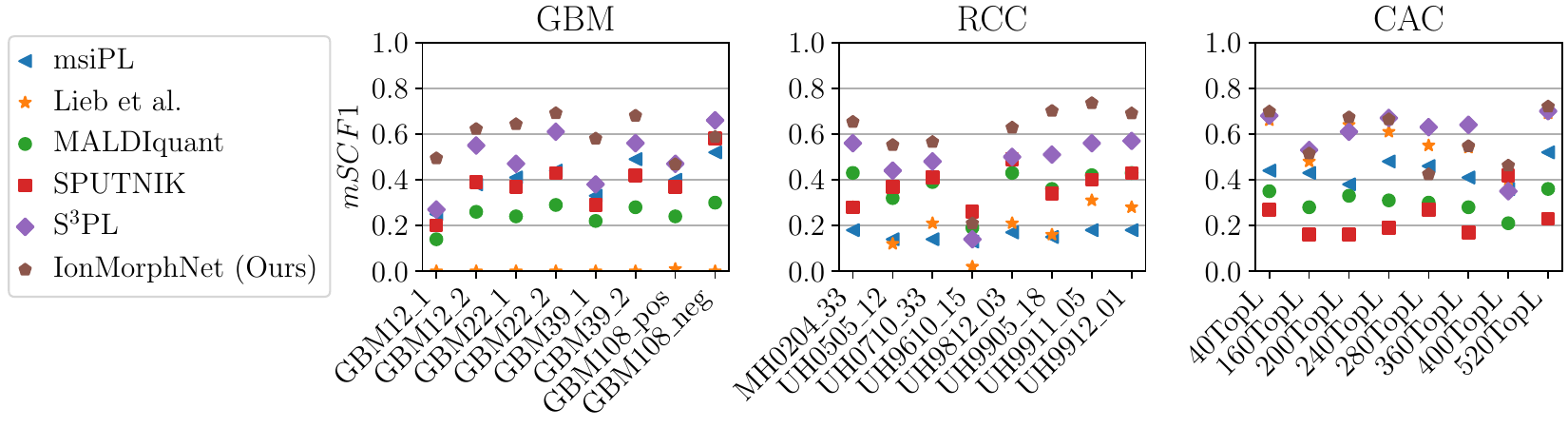}

   \caption{Detailed peak picking results of our model compared to the State of the Art \cite{abdelmoula2021peak,lieb2020peak,gibb2012maldiquant,inglese2019sputnik,weigand2026spatial} on the GBM~\cite{abdelmoula2022massnet}, RCC~\cite{bemis2019cardinalworkflows} and CAC~\cite{inglese2017deep} datasets. For further details, we refer to the benchmark paper~\cite{weigand2026spatial}. Our model picked those ion images for which the summed probability predictions for the classes \textit{structured}, \textit{localized} and \textit{negative} are the highest.}
   \label{fig:detailed_peakpicking_results}
\end{figure*}
\vspace{-4pt}
\section{Experiments}

We evaluate IonMorphNet on two complementary tasks. First, we assess its
performance for peak picking on raw mass spectrometry imaging (MSI) spectra and compare it against established peak picking methods. Second, we investigate whether IonMorphNet can serve as an effective dimensionality reduction step for downstream tumor classification using spatially aware spectral patches. In addition, we analyze the influence of several design choices through ablation studies.

\subsection{Experimental Setup}

\paragraph{Evaluation Datasets.}
For our evaluations, we use the same test datasets as described in S\textsuperscript{3}PL~\cite{weigand2026spatial} to validate our pretrained IonMorphNet. The test datasets include a mouse brain tumor model of glioblastoma (GBM) \cite{abdelmoula2022massnet}, a renal cell carcinoma (RCC) dataset \cite{bemis2019cardinalworkflows}, and a human colorectal adenocarcinoma (CAC) dataset \cite{inglese2017deep}. The datasets consist of raw profile spectra together with corresponding tissue annotations. Each dataset collection contains eight tissue sections, resulting in a total of 24 tissue sections across all datasets. Note that these datasets are not included in our labeled collection of $53$ MSI datasets.

\paragraph{Training Details.}
We employ Cross-Entropy as the loss function. All experiments are performed using weighted random sampling with a batch size of $16$, cosine learning rate schedule with warm restarts~\cite{loshchilov2016sgdr} with $8$ warmup epochs, starting with a learning rate of $3\times10^{-5}$ and a minimum learning rate of $10 ^{-6}$. Dropout \cite{srivastava2014dropout} is used with a rate of $0.05$. We train with a patience of $10$ epochs but for a maximum of $100$ epochs. We track the validation loss and choose the model with the lowest validation loss as our final model. We obtained all model architectures and pretrained weights from the timm library ~\cite{rw2019timm} and performed our experiments on a single NVIDIA A100 GPU. We use pyM2aia~\cite{cordes2024pym2aia} as an efficient batch generator for imzML files.

\subsection{Peak Picking Evaluation}

Peak picking is a standard preprocessing step for raw (profile) mass spectra, reducing data dimensionality while focusing on informative peaks and ion images. We compare our approach with current state-of-the-art peak picking methods by evaluating the spatial informativeness of the selected ion images, as recently proposed~\cite{weigand2026spatial}. To do so, our IonMorphNet is applied on each ion image of a dataset and provides a score for the degree of structure $S$ for each ion image. Given the logit vector $\mathbf{z} \in \mathbb{R}^N$, where $N=6$ denotes the number of classes, produced by our IonMorphNet, we apply softmax to obtain the predicted class probabilities $\hat{\mathbf{p}} = \text{softmax}(\mathbf{z})$. We then define the degree of structure score $S$ for a predefined set of target classes $\mathcal{T} \subseteq \{1, \dots, N\}$ as the cumulative probability mass over those classes:
\begin{equation}
    S_\mathcal{T} = \sum_{k \in \mathcal{T}} \hat{p}_k.
\end{equation}
Here, $\mathcal{T}$ may correspond to a semantically meaningful grouping of classes. For instance, we consider the classes \textit{Structured}, \textit{Negative} and \textit{Localized} as spatially informative and calculate the Score $S$ by summing the softmax probabilities of these specific classes. We will later demonstrate that this combination yields optimal peak picking results (see \cref{tab:ablation_structure}). The IonMorphNet is applied to each ion image, resulting in the generation of a list that provides a ranking of all ion images in a dataset according to their degree of spatial structure. In order to compare our approach to the state-of-the-art peak picking methods, it is necessary to select a comparable number of peaks $n$, for each tissue section. This number is provided by the evaluation protocol, and the ordered list of ion images can simply be cut off at this number $n$, leaving the top $n$ ion images with a high degree of spatial structure.

\cref{tab:averaged_results} shows the averaged mSCF1 results across the three evaluation datasets and \cref{fig:detailed_peakpicking_results} illustrates the peak picking performance for all tissue sections separately. Our approach outperforms state-of-the-art methods on $19$ out of $24$ tissue sections with an overall average mSCF1 of $59.2$~\%. In contrast to the current state-of-the-art in peak picking, we achieve these impressive results without any required parameter tuning or dataset-specific training.
Overall, our evaluation demonstrates that incorporating spatial morphology through ion image analysis enhances peak selection performance compared to purely spectral approaches.

\begin{table}[t]
\caption{Averaged peak picking results of our model compared to the State of the Art~\cite{abdelmoula2021peak,lieb2020peak,gibb2012maldiquant,inglese2019sputnik,weigand2026spatial} across the GBM~\cite{abdelmoula2022massnet}, RCC~\cite{bemis2019cardinalworkflows} and CAC~\cite{inglese2017deep} datasets.}
\label{tab:averaged_results}
\centering
\begin{tabularx}{\linewidth}{Xc}
\hline
Method                               & Avg. mSCF1 (\%) \\ \hline
Lieb et al. \cite{lieb2020peak}      & 25.7            \\
MALDIquant \cite{gibb2012maldiquant} & 30.7            \\
SPUTNIK \cite{inglese2019sputnik}    & 32.9            \\
msiPL \cite{abdelmoula2021peak}      & 33.2            \\
S\textsuperscript{3}PL \cite{weigand2026spatial} & 52.2\\
IonMorphNet (Ours)                   & \textbf{59.2}   \\ \hline
\end{tabularx}
\end{table}

\subsection{Ablation Studies}

\paragraph{Structure Class Selection.}
IonMorphNet relies on structural labels that describe characteristic spatial patterns observed in ion images. To evaluate the influence of these classes, we evaluate models using different subsets of the available structural categories. \cref{tab:ablation_structure} reports the resulting mSCF1 scores for each configuration.

\begin{table}[!h]
\caption{Ablation study on classes which are considered as spatially structured for the task of peak picking.}
\label{tab:ablation_structure}
\centering
\begin{tabularx}{\columnwidth}{YYYYY}
\toprule
Structured & Negative & Localized & Weakly Structured & Avg. mSCF1 (\%) \\
\midrule
\checkmark & \checkmark & \checkmark & & \textbf{59.2} \\
\checkmark & \checkmark & \checkmark & \checkmark & 56.0\\
\checkmark & \checkmark &  &  & 55.4\\
\checkmark & \checkmark &  & \checkmark & 53.6\\
\checkmark & & \checkmark & & 50.3\\
\checkmark & & & & 45.2 \\
\bottomrule

\end{tabularx}
\end{table}

The classes \textit{structured} and \textit{negative} are essential to include, as they contribute to the best four results. Although \textit{localized} and \textit{weakly structured} both represent subtle patterns, peak picking is most beneficial when focusing specifically on the \textit{structured}, \textit{negative}, and \textit{localized} categories.

\paragraph{Backbone Architectures.}
To investigate the impact of the feature extractor, we evaluate several
state-of-the-art CNN and ViT backbone architectures commonly used for image classification in \cref{tab:ablation_backbones}. All backbones are trained under identical conditions. Note that we chose the classes in accordance with our previously described ablation.

% Backbones
\begin{table}[t]
\caption{Comparison of different backbones for our classification model. We use the classes "structured", "localized" and "negative" to select spatially structured peaks.}
\label{tab:ablation_backbones}
\centering
\begin{tabularx}{\columnwidth}{lRYY}
\hline
\multirow{3}{*}{Backbone} & \multirow{3}{*}{Params} & Avg.& Test\\
& & mSCF1 & Acc.\\
& & (\%)& (\%)\\
\toprule
\multicolumn{4}{c}{\textit{Convolutional Neural Networks}} \\
\midrule
ConvNeXt V2-Ti \cite{woo2023convnext} & 27.9 M& \textbf{59.2} & 74.0\\
ConvNeXt V2-B \cite{woo2023convnext} & 87.7 M& 57.6 & 63.1\\
ResNet34 \cite{he2016deep} & 21.2 M& 55.2 & 70.0\\
ResNet50  \cite{he2016deep} & 23.5 M& 58.1 & 71.8\\
ResNet101 \cite{he2016deep} & 42.5 M& 58.7 & 68.1\\
ResNet152 \cite{he2016deep} & 58.1 M& 55.6 & 73.2\\
Xception  \cite{chollet2017xception} & 24.9 M& 56.4 & 71.1\\
EfficientNet-B3 \cite{tan2019efficientnet} & 10.7 M& 52.6 & 62.1\\
EfficientNet-B5 \cite{tan2019efficientnet} & 28.3 M& 55.9 & 65.2\\
\midrule
\multicolumn{4}{c}{\textit{Vision Transformers}} \\
\midrule
SwinV2-S \cite{liu2022swin} & 48.9 M& 54.7 & 71.0\\
PVTv2-B0 \cite{wang2022pvt} & 3.4 M& 54.7 & 74.3\\
PVTv2-B3 \cite{wang2022pvt} & 44.7 M& 53.0 & 72.9\\
PVTv2-B5 \cite{wang2022pvt} & 81.4 M& 53.6 & 74.3\\
DeiT-Ti \cite{touvron2021training} & 5.4 M& 55.4 & 70.9\\
DeiT-S \cite{touvron2021training} & 21.5 M& 53.8 & 67.1\\
DINOv3-S \cite{simeoni2025dinov3}& 21.4 M& 53.6 & 68.6\\
\bottomrule
\end{tabularx}
\end{table}

The ConvNeXt-V2 Tiny backbone achieves the best average mSCF1 of $59.2$ \%. It is important to note that each backbone outperforms the current State of the Art, S\textsuperscript{3}PL.

\paragraph{ImageNet Pretraining.}
We further examine whether initializing the backbone with ImageNet-pretrained weights improves performance. \cref{tab:ablation_pretraining} compares models trained from random initialization with models initialized using pretrained weights.
The results show that ImageNet pretraining improves the performance in terms of mSCF1 in most cases. This suggests that generic visual features learned from natural images can transfer to ion image analysis despite the domain difference.

% Pretrained vs. from Scratch
\begin{table}[]
\centering
\caption{Comparison of \textit{Avg. mSCF1} results for ImageNet \cite{deng2009imagenet} pretrained backbones and training from scratch.}
\label{tab:ablation_pretraining}
\begin{tabularx}{\columnwidth}{lYY}
\toprule
\multirow{2}{*}{Backbone} & \multicolumn{2}{c}{Avg. mSCF1 (\%)} \\ \cmidrule(lr){2-3}
& Pretrained & from Scratch \\
\midrule
ConvNeXt V2-Ti \cite{woo2023convnext} & \textbf{59.2} & 48.4 \\
ResNet101 \cite{he2016deep} & \textbf{58.7} & 50.7 \\
ResNet50 \cite{he2016deep} & \textbf{58.1} & 52.9 \\
Xception \cite{chollet2017xception} & 56.4 & \textbf{56.6} \\
SwinV2-S \cite{liu2022swin} & \textbf{54.7} & 53.1 \\
DeiT-Ti \cite{touvron2021training} & \textbf{55.4} & 54.3 \\
\bottomrule
\end{tabularx}
\end{table}

\subsection{Downstream Tumor Classification}
Beyond peak picking, we evaluate whether the channels selected by
IonMorphNet improve tumor classification as a downstream MSI task. The goal of this experiment is not to introduce a new classification architecture, but to demonstrate that the learned channel reduction preserves — and can even enhance — discriminative information relevant for clinically meaningful tasks.

\paragraph{Spatio-Spectral Patch Sampling.}
Tumor classification in MSI is commonly performed by classifying each spectrum independently using the raw spectrum~\cite{behrmann2018deep, kriegsmann2020mass, kanter2023classification}. However, this approach ignores the spatial context that is provided in MSI datasets. In Hyperspectral Image Classification, spatially aware models such as 3D CNNs~\cite{paoletti2019deep} can be used due to the comparably low number of channels ($100-200$). Since the number of model parameters of a 3D CNN increases drastically with more than $10,000$ channels, their application is impractical for unprocessed MSI data.

We address these limitations by applying our proposed peak picking IonMorphNet for dimensionality reduction prior to tumor classification. For each dataset collection, we perform peak picking on all eight tissue sections and select every m/z value that was picked for any of the eight tissue sections. This ensures the same channel count across all tissue sections. As our IonMorphNet produces an ordered list of spatially structured ion images for each tissue section, we selected the top $n$ picked peaks so that the final peak number is close to those of the state-of-the-art methods. We divide each dataset collection into three training, two validation and three testing tissue sections. We use IonMorphNet with a ConvNeXt V2 - Tiny Backbone, as it achieved the best average mSCF1 in our ablation study (see \cref{tab:ablation_backbones}).

\begin{figure*}[!t]
  \centering
   \includegraphics[width=1.0\linewidth]{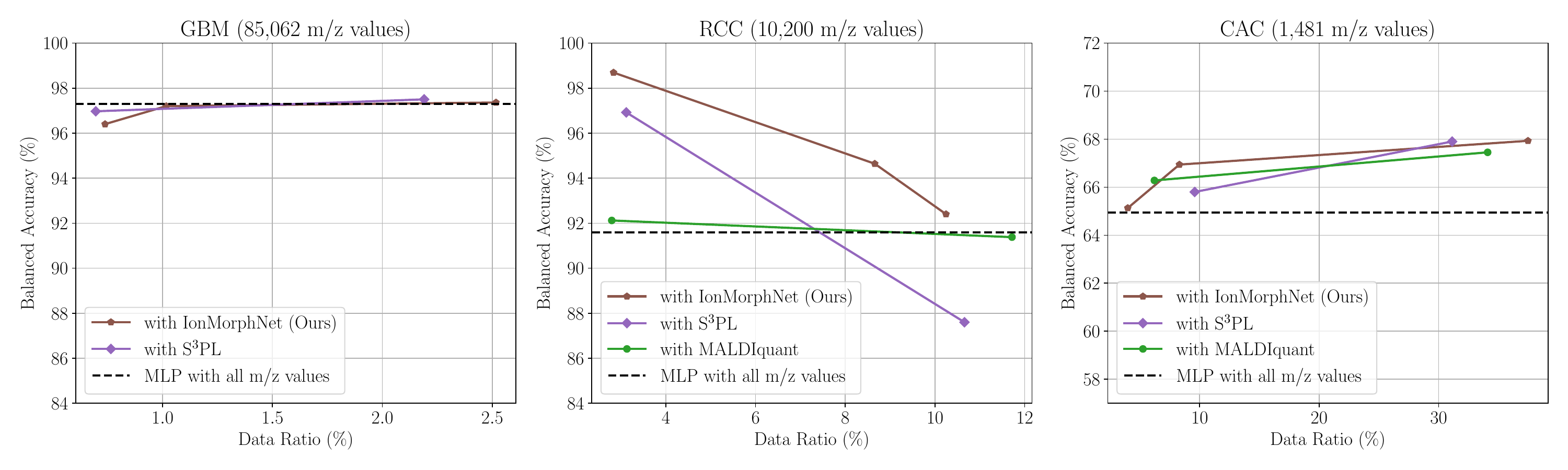}

   \caption{Tumor classification results of a 3D CNN~\cite{paoletti2019deep} on peak picked datasets \cite{abdelmoula2022massnet, bemis2019cardinalworkflows, inglese2017deep} with a spectral patch size $p=11$. We compare different peak picking methods and data ratios of the entire spectra. Note that MALDIquant~\cite{gibb2012maldiquant} was unable to generate a unified peak list for the GBM~\cite{abdelmoula2022massnet} dataset due to limitations in memory capacity.}
   \label{fig:3DCNN_results}
\end{figure*}

\paragraph{Classification Setup.}
We extract a spectral cube of size ($p \times p \times C$) around each spatial coordinate and assign it the label of the central spectrum. Here, $p$ refers to the spatial dimension of the spectral cube and $C$ to the channel dimension, which is the number of peak picked channels. 
Tumor classification is performed using a 3D convolutional neural network that processes the resulting spatio-spectral patches. For comparison, we also train an MLP for single spectrum classification of unprocessed mass spectra. For each test tissue section, the Balanced Accuracy is calculated separately and we report the mean Balanced Accuracy across all test tissue sections.

Cross-Entropy is utilized as the loss function. All experiments are performed using weighted random sampling with a batch size of $16$ and a learning rate of $10 ^{-4}$. The 3D CNN\cite{paoletti2019deep} is trained for three epochs, as it typically converges after one epoch. The MLP is trained for $100$ epochs. The validation loss is used to select the best model. Final evaluation is performed on the test tissue sections. The selected 3D CNN architecture represents the current state-of-the-art method on the Hyperspectral Imaging Benchmark~\cite{frank2023hyperspectral}. Results are averaged across five seeds. All experiments are performed on a single  NVIDIA A100 GPU.

\paragraph{Results.}
\cref{fig:3DCNN_results} shows the classification performance of the 3D CNN and the MLP on the test tissue sections of the GBM~\cite{abdelmoula2022massnet}, RCC~\cite{bemis2019cardinalworkflows} and CAC~\cite{inglese2017deep} datasets, respectively. For the 3D CNN, different channel ratios resulting from peak picking are used. The horizontal black lines represent the Balanced Accuracy of the MLP trained on single unprocessed spectra. Note, that every peak picking method selects a specific number of peaks, which is parameter dependent. Therefore, we can only report results for similar but not identical data ratios.

For the GBM dataset, the MLP result of $97.4 \%$  Balanced Accuracy is matched by the 3D CNN on peak picked spectra from S\textsuperscript{3}PL and our approach, where only about $1 \% - 2 \%$ of the entire spectrum is utilized. MALDIquant~\cite{gibb2012maldiquant} was unable to generate a unified peak list for the GBM dataset due to limitations in memory capacity. When using less than $1 \%$ of the full spectrum, the performance declines.% to $96.0 \%$ Balanced Accuracy. 
For the RCC dataset, channel reduction by peak picking with our IonMorphNet consistently improves results over the MLP result and compared to other peak picking methods with similar amounts of data. Interestingly, the Balanced Accuracy increases with fewer data, resulting in the best improvement of $+7.3 \%$ Balanced Accuracy over the single spectrum MLP result with approximately $3 \%$ of the channels. For the CAC dataset, all peak picking methods enable the 3D CNN equally well to enhance Balanced Accuracy by up to $+3 \%$. Note that we selected higher data ratios for this dataset, because the unprocessed spectra already consists of relatively few channels ($1,481$) compared to the GBM ($85,062$ channels) and RCC ($10,200$ channels) dataset.

These results suggest that peak channels identified through spatial
morphology analysis retain biologically meaningful information that is beneficial for tumor classification as a downstream task.

\paragraph{Patch Size Ablation.}
Finally, we evaluate the influence of the spatial patch size $p$ used during classification. Different patch sizes control the spatial context available to the classifier.

\begin{table}[t]
\centering
\caption{Ablation study on different patch sizes $p$ for the spatio-spectral sampling for tumor classification using a 3D CNN~\cite{paoletti2019deep}.}
\label{tab:ablation_patchsize}
\resizebox{\columnwidth}{!}{%
\begin{tabular}{ccccccc}
\toprule
\multirow{2}{*}{patch size} & \multicolumn{2}{c}{GBM} & \multicolumn{2}{c}{RCC} & \multicolumn{2}{c}{CAC} \\ \cmidrule(lr){2-3} \cmidrule(lr){4-5} \cmidrule(lr){6-7}
 & 0.75 \% & 2.5 \% & 2.9 \% & 10.2 \% & 8.3 \% & 37.5 \% \\ 
 \midrule
11 & \textbf{96.4} & \textbf{97.4} & 98.7 & 94.6  & \textbf{66.9} & 67.9 \\
13 & 93.3 & 97.1 & 97.0 & 95.1  & 65.7 & 69.3 \\
15 & 92.5 & 95.4 & \textbf{98.9} & \textbf{95.8} & 64.3 & \textbf{69.5} \\
\bottomrule
\end{tabular}}
\end{table}

\cref{tab:ablation_patchsize} summarizes the resulting performance for several patch size configurations. The classification performance benefits from smaller patch sizes for the GBM dataset which suggests that local spatial context provides the best contextual information in this case. For the RCC dataset, the performance is stable across various sizes for $p$. Note that the 3D CNN architecture requires a minimum patch size $p=11$.
\section{Conclusion}
We present IonMorphNet, the first generalizable ion image encoder for ion image morphology assessment in MSI. Through a diverse dataset collection of $53$ public MSI datasets, IonMorphNet enables generalizable and hyperparameter free peak picking, outperforming state-of-the-art methods by $+7 \%$ Avg. mSCF1. Furthermore, we demonstrate that spatially informative channel reduction enables 3D CNNs for spatio-spectral tumor classification, improving classification performance compared to conventional pixel-wise approaches.
\vspace{-4pt}
\section*{Acknowledgments}
This project has been partially funded by the Ministry of Science, Research and Arts Baden-Württemberg in the project Perpharmance (BW6-07) and the German Research Foundation (INST874/9-1).
{
    \small
    \bibliographystyle{ieeenat_fullname}
    \bibliography{main}
}


\end{document}